\documentclass[conference]{IEEEtran}
\IEEEoverridecommandlockouts
\usepackage{cite}
\usepackage{amsmath,amssymb,amsfonts}
\usepackage{algorithmic}
\usepackage{graphicx}
\usepackage{textcomp}
\usepackage{xcolor}
\usepackage{hyperref}
\usepackage{breakurl}

\usepackage[T1]{fontenc}  
\usepackage[utf8]{inputenc}

\def\BibTeX{{\rm B\kern-.05em{\sc i\kern-.025em b}\kern-.08em
    T\kern-.1667em\lower.7ex\hbox{E}\kern-.125emX}}
\begin{document}

\title{Machine Learning Model Attribution Challenge
\thanks{©2023 The MITRE Corporation. ALL RIGHTS RESERVED\\
Approved for public release. Distribution unlimited 22-03442-4.}
}

\author{\IEEEauthorblockN{
Elizabeth Merkhofer\IEEEauthorrefmark{1},
Deepesh Chaudhari\IEEEauthorrefmark{2},
Hyrum S. Anderson\IEEEauthorrefmark{3}, \\
Keith Manville\IEEEauthorrefmark{1},
Lily Wong\IEEEauthorrefmark{1},
João Gante\IEEEauthorrefmark{4}
\\
\IEEEauthorrefmark{1}MITRE Corporation \textit{\{emerkhofer, kmanville, lwong\}@mitre.org}, \\
\IEEEauthorrefmark{2}Plaintext Group \textit{deepesh@plaintextgroup.com}, \\
\IEEEauthorrefmark{3}Robust Intelligence \textit{hyrum@robustintelligence.com}, \\
\IEEEauthorrefmark{4}Hugging Face \textit{joao@huggingface.co}
}}
\maketitle

\begin{abstract}

We present the findings of the Machine Learning Model Attribution Challenge.
Fine-tuned machine learning models may derive from other trained models without obvious attribution characteristics.
In this challenge, participants identify the publicly-available base models that underlie a set of anonymous, fine-tuned large language models (LLMs) using only textual output of the models.
Contestants aim to correctly attribute the most fine-tuned models, with ties broken in the favor of contestants whose solutions use fewer calls to the fine-tuned models' API.
The most successful approaches were manual, as participants observed similarities between model outputs and developed attribution heuristics based on public documentation of the base models, though several teams also submitted automated, statistical solutions.
\end{abstract}


\section{Introduction}

The announcement of GPT-2, a large language model (LLM) that could generate text of ``unprecedented quality,'' brought new concerns to the forefront of responsible AI in 2019 \cite{gptblog}. 
Ongoing algorithmic advances and increases in compute capacity have enabled big tech companies to train multi-billion parameter neural networks on web scale collections of text.
The resulting models write so fluently that even humans trained to detect machine-generated content can be fooled about 30\% of the time \cite{ippolito-etal-2020-automatic}.
Meanwhile, the culture in the natural language processing (NLP) community has moved toward open sourcing these `foundation models' with permissive licenses, allowing much smaller and less-resourced entities to deploy them, often after further training them to specialize them to a different domain or task (fine-tuning)\cite{foundationmodels}.
Since publicly released, trained LLMs can be fine-tuned at a relatively low cost and generate specialized synthetic text at scale, they may empower more actors to engage in malicious uses, for example through cheaper and more effective disinformation \cite{deepmindethical, infop}.
It is unclear how widely LLMs are used for deceptive purposes today.

Possible mitigations for this new threat include ensuring that synthetic text can be identified as non-human and its origins traced \cite{infop}.
Incorporating a watermark into LLM output has been proposed as a safeguard to privacy and intellectual property \cite{fang-etal-2017-generating, aaronson_2022}, but is not widely practiced. 
This competition asks what traces of provenance can be gleaned from synthetic text that lacks explicit marking.
To our knowledge, no generalized forensic process exists to trace textual output from customized models back to the base model.
Attribution techniques could represent a powerful tool for regulation, tracking, and remediation where LLMs are misused.

The Machine Learning Model Attribution Challenge calls upon contestants to develop creative solutions to uncover model provenance. 
Contestants interact with a set of fine-tuned models via a text generation API, attributing each back to a known set of LLMs (`base models'). 
Building out forensic capabilities and establishing the difficulty of model attribution is a step toward assured use of LLMs and artificial intelligence in general.

In this paper, we describe the competition and summarize new methods of attribution proposed by the participants.

\section{Competition details}

MLMAC launched at DEFCON 30’s AI Village with promotion through the Village and sponsor social media sites.
The first round, described in this paper, lasted for five weeks, from August 12, 2022, anywhere on Earth (AoE), to September 16, 2022, AoE. 
Information for participants, including terms and conditions and sample code for interacting with the competition API, is hosted at \href{https://mlmac.io}{https://mlmac.io}.

A further round, not detailed in this paper, was hosted on Kaggle in November and December 2022.

\subsection{Scenario}

We consider a scenario in which an adversary has deployed a language model for text generation via an API where users can input a prompt and receive a textual continuation from the model.
Because of the prohibitive cost of training LLMs from scratch, it is likely the model is based on another model, using fine-tuning to tailor the output to a particular use case.
There are many base models that have been open sourced for research by large corporations, like GPT models by OpenAI, while other models are intended to remain proprietary.
Knowing the provenance of the model may reveal clues about the adversary's affiliations.
It would also allow a better understanding of the fine-tuned model, e.g. what data was included in the base model training and the size, architecture and vulnerabilities of the model.
Attributing the model's origins may also uncover IP theft, if the adversary's model is based on a `stolen' proprietary model, or abuse, if an open source model being used outside the model's licensing agreement or terms of service.
Challenge participants attempt to discover the origin of the fine-tuned model by interacting with it via the adversary's natural language generation API, but aim to keep their total number of calls low to avoid detection.
They have full access to a set of documented `base models' that likely underlie the adversary's model.

In our scenario, base models have not been watermarked, and no particular effort has been made by the adversary to obfuscate the fine-tuned model’s provenance.

\subsection{Participation}

A competition API is provided for participants to interact with the fine-tuned and base models.
Participants interact with fine-tuned models using integer IDs and base models using the model's ID from the Hugging Face model hub\footnote{\url{https://huggingface.co/models}}.
The API provided by the MLMAC team wraps Hugging Face's Text Generation inference API, allowing participants to generate text from the models starting with a conditioning string. 
All other decoding settings are fixed so that continuations are generated using sampling with a temperature of 3.0. 
This reflects the threat scenario where participants have limited access to an adversary's fine-tuned model.
Calls to the fine-tuned models are counted, as the number of calls factors into the competition ranking. 

To mimic a forensic scenario where participants have open access to the base models, use of base models is not constrained.
Calls to the base models through the MLMAC API are not counted or factored into the competition rankings. 
Participants may also choose to interact with the base models directly by downloading them from the Hugging Face hub and using the transformers library, e.g. to obtain model probabilities or test different sampling parameters.
Model cards and other documentation on the Internet provide participants with more information about the base models.

Solutions are submitted in the form of \textit{(fine-tuned model, base model)} pairs.
Participants are informed that there may not be a one-to-one mapping between base and fine-tuned models: there may be extraneous base models, multiple fine-tuned models with the same base, and/or fine-tuned models that are not derived from any of the provided base models.
An example solution is shown in Table \ref{tab:example}.

\begin{table}[]
    \centering
        \caption{Sample solution demonstrating an unused base model $E$ and a pairing with $None$. }
    \begin{tabular}{| c||c|c|c|c|c|c|} 
    \hline
         &  model 1 & model 2 & model 3 & model 4 & model 5 \\ \hline
         base model A &X& & & & \\ \hline
         base model B & &X& & & \\ \hline
         base model C & & & &X& \\ \hline
         base model D & & & & &X \\ \hline
         base model E & & & & & \\ \hline
         None         & & &X& & \\ \hline
    \end{tabular}

    \label{tab:example}
\end{table}



\begin{table}[]
    \centering
    \caption{Participants and results of MLMAC.}
    \begin{tabular}{|c|c|r|c|} 
    \hline
        Username & \# Correct & \# Queries & Student \\ \hline
    Pranjal2041& 7 & 1212 & X \\
    YoulongDing& 6 & 168 & X \\
    Jordine& 6 & 244 & X \\
    FarhanDhanani& 6 & 1084 & X \\
    \textit{MLMAC Team Baseline}& 6 & 500000 & \\
    sheetal57& 5 & 604 &  \\
    curranjanssens& 4 & 516 & X \\
    ogozuacik& 4 & 13825 &  \\
    nick-jia& 3 & 12 & X \\
    JosephTLucas& 3 & 843 &  \\
    ambrishrawat& 3 & 1725 &  \\
    oleszko& 2 & 11 &  \\
    ri638& 1 & 0 & X \\
    Saifulislamsalim79& 0 & 2 & X \\
        \hline
    \end{tabular}

    \label{tab:leaderboard}
\end{table}

\begin{table*}[t]
\centering
\caption{Documented Participant Approaches}
\begin{tabular}{|p{0.07\linewidth}|p{0.3\linewidth}|p{0.53\linewidth}|}
\hline
          & Approach                                       & Specific approach                                                        \\ \hline
\hfil Manual    & Reasoning about base model training data                       & Temporal range \cite{pranjalblog};
    Domain \cite{ding, jordine, pranjalblog};
    Language \cite{ding, jordine, pranjalblog}; 
    Model-specific special tokens \cite{pranjalblog}; Length \cite{jordine, pranjalblog} \\
          & Similar output to base model, given same query & Gibberish/repetition \cite{ding, jordine};
          Same continuation \cite{jordine}                        \\
          & Observations about the API                     & Time to load and API failures \cite{ding, jordine}                                   \\ \hline
\hfil Automatic & Similar output to base model, given same query 
    & Edit distance \cite{lucas_2022}; 
    Machine translation metrics \cite{farhanarxiv};   
    Next character (baseline) \\
          & Similar vocabulary use                         & Term-document vector similarity \cite{farhanarxiv}                                 \\
          & Model of base model output                     & Neural network attribution model    \cite{farhanarxiv}                 
          \label{tab:approaches}\\  \hline

\end{tabular}
\end{table*}

\begin{table*}[]
\centering
\caption{Competition Solution. 
Models are identified by their names in the Hugging Face hub.}
\begin{tabular}{|p{0.05\linewidth}|p{0.45\linewidth}|p{0.25\linewidth}|p{0.08\linewidth}|}
\hline
 Model ID & Fine-tuned Model                                              & Base Model                    &  \# Correct Attributions \\ \hline
\hfil0        & LACAI/DialoGPT-large-PFG                                      & microsoft/DialoGPT-large      & \hfil4                       \\ \hline
\hfil1        & BramVanroy/gpt-neo-125M\_finetuned-tolkien                    & EleutherAI/gpt-neo-125M       & \hfil4                       \\ \hline
\hfil2        & MultiTrickFox/bloom-2b5\_Zen                                  & bigscience/bloom-2b5          & \hfil7                       \\ \hline
\hfil3        & mrm8488/GPT-2-finetuned-covid-bio-medrxiv                     & gpt2                          & \hfil0                       \\ \hline
\hfil4        & projectaligned/gpt2-xl-reddit-writingprompts-behavior-cloning & gpt2-xl                       & \hfil5                       \\ \hline
\hfil5        & aliosm/ComVE-distilgpt2                                       & distilgpt2                    & \hfil2                       \\ \hline
\hfil6        & google/reformer-crime-and-punishment                          & \textit{None}                 & \hfil1                       \\ \hline
\hfil7        & Salesforce/codegen-350M-mono                                  & Salesforce/codegen-350M-multi & \hfil9                       \\ \hline
\hfil8        & wvangils/BLOOM-350m-Beatles-Lyrics-finetuned-newlyrics        & bigscience/bloom-350m         & \hfil1                       \\ \hline
\hfil9        & hakurei/lit-6B                                                & EleutherAI/gpt-j-6B           & \hfil4                       \\ \hline
\hfil10       & textattack/xlnet-base-cased-rotten-tomatoes                   & xlnet-base-cased              & \hfil9                       \\ \hline
\hfil11       & lvwerra/gpt2-imdb                                             & gpt2                          & \hfil4                       \\ \hline
\hfil-        & \textit{None}                                                 & facebook/opt-350m             & \hfil-                       \\ \hline
\end{tabular}
\label{tab:solution}
\end{table*}

\subsection{Evaluation}
Solutions are evaluated according to the following rank-ordered criteria:
\begin{enumerate}
\item \textbf{Number of correct pairs.}
\item \textbf{Fewest queries to fine-tuned models}: ties are broken by selecting the contestant who used the fewest API queries to interact with the fine-tuned models. (Queries to the base models are not counted.)
\item \textbf{Earliest submission time}: any subsequent ties are be broken by selecting the contestant whose final submission was earliest.
\end{enumerate}

Prizes were awarded to winning solutions: USD \$3,000 and USD \$1,500 for second place, respectively. In addition, the three highest-ranking student submissions were awarded additional support to attend either CAMLIS 2022 or SaTML 2023 in the form of conference admission and USD \$1,500 to offset travel and accommodation costs.
Participants were required to publish their solution to qualify for prizes. These could take any form, such as blog post or pre-print paper. Schmidt Futures and Mercatus Center generously provided funding for logistics, prizes and travel grants.

\section{Results}

Thirteen teams submitted solutions to MLMAC. Their accuracy and number of queries are reported in Table \ref{tab:leaderboard}.
Participants' approaches are broadly characterized in Table \ref{tab:approaches}.
The most successful approaches are accurate about half the time.
The top-scoring team correctly identified seven base models. Three additional teams got six correct, with their submissions differentiated by the number of calls to the fine-tuned model APIs.
The top two teams were awarded the cash prizes, and the top three teams, all students, received the student travel awards. 
The top four teams and participant JosephTLucas shared details of their approach through blog posts, notes and an arXiv paper \cite{ding, farhanarxiv, jordine, lucas_2022, pranjalblog}. 
The number of API calls was relatively low among most submissions, from hundreds to low thousands of calls. 
While several teams submitted answers with an extremely low number of calls, none of these groups shared any information about their approach.

The solution to the challenge is presented in Table \ref{tab:solution}. Note that \texttt{gpt2} is the base model of two fine-tuned models, one fine-tuned model should be attributed to none of the base models in the set, and the list of base models provided to the participants included \texttt{facebook/opt350m} though there was no fine-tuned version in the challenge set. The number of correct attributions per model ranges from zero to nine of the thirteen participating groups. 
The two pairs that are most often correctly identify have particularly distinctive output: fine-tuned model 7 and its base model both output computer code, and fine-tuned model 10 and its base model both produce repetitive text.

\subsection{Manual Approaches}

The top three teams all used manual approaches with no automatic component. The participants hand wrote and/or curated prompts to test model output for particular clues.
These prompts were mostly well-formed language, not gibberish or random strings. 
Then they manually reviewed the outputs and reasoned about them.
Their clues fell under three categories: differences between base model training data, similarity in output between models, and observations about the competition API.

The participants reasoned about the fine-tuned models by searching for differences they could trace to the training data of the base model. 
The training data used for most base models is documented online in the model card. 
Participants assume that characteristics of this data will persist after fine-tuning or are consistent with any fine-tuning data.
Several observed that the temporal range of base model training data varied, and looked for clues using prompts about the current president or COVID-19.
Several models were trained on particular domains, such as code or dialogue, and participants reasoned that fine-tuned models would share the same domain.
Base models are trained on different groups of languages or monolingual English, so participants examined how the models continued prompts in other languages, such as Chinese and Indic languages.
One participant attempted to insert model-specific tokens into the prompt, such as those used between different text sequences during training, to observe model behavior.
Participants documented additional conclusions about fine-tuning data that they were not able to apply to the competition, like that one was trained on Beatles lyrics.

Participants also observed similarities in output between base models and fine-tuned models. One base model and one fine-tuned model tended to output repetitive gibberish. In addition, several fine-tuned and base model pairs were observed to have the same continuation for some natural language prompts. This was even observed between different sizes of the same model architecture.

Finally, two participants used observations about the competition API, specifically that two models took much longer to load and failed to return a response more often. They attributed these fine-tuned models to the largest base models.

\subsection{Automatic Approaches}

Two participants report using automatic approaches. 
Both used diverse prompts to increase the odds of finding continuations where the fine-tuned model was very similar to the base model, one by creating random strings and the other by selecting a set of inputs from public NLP datasets.
These participants measured similarity between outputs for the same prompt using similarity metrics, such as those from machine translation, and and term vectors like those used in information retrieval. 
Ultimately, the fourth-place participant trained a neural network model to discriminate between the outputs of base models, then submitted their attribution based on the model's predictions on a set of fine-tuned model outputs.

Our leaderboard includes a baseline solution by the MLMAC team, correctly attributing six models. 
This automatic solution generated random strings and obtained 100 continuations from each model for each using beam search (beam search decoding would not be possible under the competition API). 
Each fine-tuned model was attributed to the base model that had the most similar distribution of first characters of the continuations. 
The total number of strings generated to support this approach was much higher than any participant's solution.
The baseline solution was not shared with participants, but rather used to set expectations of the difficulty of the challenge.


\section{Conclusions}

Language model attribution techniques are still in their infancy. 
Our most successful participants used manual approaches that relied on hand-crafting prompts and reasoning about public information shared about the base models.
These techniques will not scale to a larger number of possible base models, especially base models that share training data, and don't establish a `fingerprint' that could positively attribute a stolen model when using a process of elimination is not possible.
It's possible, however, that the features used in participants' manual analyses could inform attribution algorithms in the future.
Furthermore, automated approaches that measure output similarity between base and fine-tuned models from the same prompt yielded some success.
Future approaches could build on this work by learning input prompts that produce more unique outputs in a given model.
The proposed solutions so far use a very modest number of API calls; it remains to be seen if their accuracy would improve as API calls increased, especially for automatic approaches.

This competition has established a baseline for model attribution for non-watermarked models.
Attribution techniques represent a powerful companion tool for regulation, tracking, and remediation. Even if security measures like watermarking become common practice, attribution techniques could serve as forensic tools as adversaries try to evade them. 

\section*{Acknowledgment}

The authors would like to thank the following colleagues for their role in shaping this competition: Jonathan Broadbent, Christina Liaghati, Ram Shankar Siva Kumar, and Yonadav Shavit. The authors would also like to thank Schmidt Futures and Mercatus Center for providing funding for the logistics, prizes and travel grants.

\bibliographystyle{IEEEtran}
\bibliography{lizcites}

\end{document}